# Uncertainty Principle based optimization; new metaheuristics framework

Mojtaba Moattari
*Signal Analysis Laboratory*
*Bio-electrical engineering school*
*Biomedical Engineering Department of Amirkabir University of Technology, Tehran, Iran*
*moatary@aut.ac.ir*

Mohammad Hassan Moradi
*Signal Analysis Laboratory*
*Bio-electrical engineering school*
*Biomedical Engineering Department of Amirkabir University of Technology, Tehran, Iran*
*mhmoradi@aut.ac.ir*

Emad Roshandel
*Eram Sanat Mooj Gostar*
*Shiraz, Iran*
*e.roshandel@esmg.co.ir*

**Abstract**

To more flexibly balance between exploration and exploitation, a new meta-heuristic method based on Uncertainty Principle concepts is proposed in this paper. UP is is proved effective in multiple branches of science. In the branch of quantum mechanics, canonically conjugate observables such as position and momentum cannot both be distinctly determined in any quantum state. In the same manner, the branch of Spectral filtering design implies that a nonzero function and its Fourier transform cannot both be sharply localized. After delving into such concepts on Uncertainty Principle and their variations in quantum physics, Fourier analysis, and wavelet design, the proposed framework is described in terms of algorithm and flowchart. Our proposed optimizer's idea is based on an inherent uncertainty in performing local search versus global solution search. A set of compatible metrics for each part of the framework is proposed to derive preferred form of algorithm. Evaluations and comparisons at the end of paper show competency and distinct capability of the algorithm over some of the well-known and recently proposed metaheuristics.

# 1 Introduction

One branch of soft computing is set of inexact methods in order to deal with tasks that are computationally hard to solve. Nature-inspired metaheuristics optimization methods are one of the subsets of soft computing, able to find the optimum solutions in a search space with unknown distributions.

Various optimization approaches have been proposed for different types of search space. For instance, Genetic Algorithm (GA) and Simulated Annealing (SA) are categorized in the class of fast algorithms in highly stochastic search spaces with scarce and wide convex localities [4]. However, due to the greedy properties of SA, it cannot deal with multimodal and narrowly convex compartments in a search space [3]. For a smooth and continuous search space, Tabu Search (TS) is considered as an appropriate method. TS sets regions of search space as taboo not to relocate solutions there anymore[5]. These algorithms have paved their ways in variety of engineering applications where their search space do not have straightforward characteristics like differntiability, convexity and smoothness, where needed for the stochastic approximators work on them. One of the challenging examples in this context is Edge Computing (EC), shifting computations for network parameters optimization to the network edges. Suffering from optimization intractability, a large number of optimization metaheiristics have been introduced or developed for EC [29-32]. Having recent substantial progress in network cost reduction and Packet Delivery Ratio (PDR) improvement, Edge Computing demands new helpful optimizers being able to seek for parameters more efficiently more than ever. The proposed physic-inspired metaheuristic in this paper is an effort to provide solution for engineering applications like EC, with a lot of heterogenous systems interacting in it involving a lot of physical laws of nature. In the next paragraphs, preliminaries for designing a new metaheuristics is established, to make it useful for a unified physical characteristic inherent in all physical interacting systems, especially emergent more in interactions, which one instance is Edges of communicating devices. The physical law under elaboration is Hiesenberg Uncertainty Principle.

A linear state space is convex and continuous, where deterministic gradient update converges to optimal point. In contrary, engineering systems may be generally nonlinear with locally linear state space over each operating point. Therefore, they demand memetic algorithms, that more act based on the memory of their context. Memetic Algorithms are a group of metaheuristic algorithms equipped with new local searching techniques to reduce the likelihood of premature convergence [6]. GA as a global solution seeker benefits from a local search engine named Nelder Mead Simplex, a powerful method which is called GA-NMD is developed [7]. Gradient Descent Hybrid of particle swarm optimization (PSO) called G-PSO and Nelder mead simplex-PSO (NM-PSO) are other examples of memetic algorithms which hybridize the conventional PSO [8], [9]. Memetics may not necessarily contain exact methods like gradient methods. The optimization methods usually use other local intensification techniques when gradient information in search locations is not available. For instance, Kransnogor et.al proposed a memetic algorithm with a self-adaptive local search [10]. This algorithm is a hybrid of GA as a global search and Monte Carlo method as a local searcher. When the population converges, the algorithm may rather explore and diversify the search space. Vice versa, the algorithm does local searching when the solutions are diverse enough.

Yet there is no unified framework adaptively capable of fine-tuning global and local search rate, able to discern when and which regions to choose for search. This issue is only cured in hybrid or memetic algorithms. This paper proposes a new physics-inspired algorithm not only to address the mentioned issue but to also deal with a unified framework and not modular hybridization. Such an approach leads to the designation of flexible architectures for attaining the best interplay among exploration and exploitation especially because the limitations caused by the fixed structure of each algorithm may not restrict. The framework decides when and where to exlplore and exploit. The suitability of proposed approach is handling both questions, with the help of metrics that measure solution update in specific region of search. For this objective, this paper proposes an algorithm inspired by one of the fundamental principles of nature, i.e. Uncertainty Principle (UP).

In quantum mechanics, Heisenberg Uncertainty Principle (HUP) is a mathematical inequality that asserts a limit to precision in measurements of certain quantities, such as position and momentum, especially when the measurements take place simultaneously [11]. In other words, quantities like speed and position of a photon cannot be measured together with high precision [12]. One instance is the act of measurement of photon location which affects the system and, consequently, the momentum of the measurand. Therefore, the more precise the photon location is measured, the less certain one gets about the truth of measured momentum. Also, this holds contrariwise. The more certain speed gets measured, the less certain it gets to measure position [13]. In upcoming sections, the paper explains more about this principle and its applications in various branches of science. This concept deeply meshes with the way optimization takes place. Metaheuristic optimizations have got so popular due to the success of their general approaches in dealing with uncertainty of choosing local versus global search. Another challenge is to select a better region over multiple search regions each time. So, by approaching a search space through this fundamental viewpoint, it is easier to create efficient optimization algorithms. To prove that UP is more than an irrelevent subject to optimization, various applications of that in different branches of science have been explained. HUP can be evident in any type of bottleneck problems having restricted information transfer channel. To elaborate, each trial can only give information about the specific quantity or aspect of a problem.

For the case of continuous optimization, either sampled solutions can be measured randomly over the whole of the solution space, or are sampled from a reduced region. The first case leads to more certainty in global spaces, while the second one leads to more certainty over locality of a solution. However, both degrees of certainties for being the global optimum are necessary and complimentary for the optimization process to be accomplished. The certainty of solution fitness for a local space is more than certainty for larger space or global solution. In the HUP, the certainty of specifying a local solution affects the certainty of the solution for a more global region. Thus, by defining these imaginary quantities, which are local-ness and global-ness of a solution, a measure of each one may affect the other. UP lies within locality and global-ness the same way speed of a photon relates to its location in quantum mechanics.

The main idea of this paper is not necessarily abstracted from the HUP in quantum mechanics. Generally, abstraction can be taken from cases in any problem where certainty about a phenomenon affects uncertainty in another phenomenon. Main characters of the proposed framework are as below:

- Decide where to explore or exploit.
- Select intensification or diversification when necessary
- A flexible framework for various problem definitions and search spaces
- A probabilistic framework with mostly randomly selected parameters
- A metric-based framework, easy to revise and develop.

This paper is organized as follows. In section 2, the UP is described in detail and also three subjects have been chosen to explain its usage. Afterwards, a deduction is summed up to realize what kind of problems can be approached through the UP. Section 3 puts forth a general framework and algorithm of UP based optimization as the main contribution of this paper. The metric tables for three-phase of region selector, uncertainty metrics, and update methods are introduced and the mindset behind that is explained. Moreover, a specific set from introduced metrics is used to design preferred algorithms for the proposed framework and the flowchart is shown up. Section 4 evaluates the algorithms and compares them with mentioned memetic algorithms and also recently proposed algorithms in metaheuristics.

# 2 Applications of uncertainty

In this section, applications of UP are checked out to generalize this concept for an optimization problem. Then, a new model for metaheuristic optimization is proposed based on this principle.

## 2.1 The UP in quantum physics

Any kind of mathematical inequality among two conjugate variables is the direct result of the UP. For example, In modern physics, the more precise the position of some particle be, the less precise its momentum has to be [13]. General formulation of uncertainty inequality is realized through standard deviations of conjugate particles:

$$\sigma_x \sigma_p \geq \frac{h}{2} \tag{1}$$

where $\sigma_p$ is the standard deviation of the momentum of the particle and $\sigma_x$ is the standard deviation of its position. This product cannot trespass half of the Planck constant. The inequality directly originates from the Fourier analysis of wave function. To be more precise, in small scales, the momentum of particles can be derived by the Fourier transform of particle locations in the wave function. This happens because particles in small scales manifest wave-like behavior. This kind of behavior allows waves with different frequencies to superpose. Superposition of waves makes them localized in special positions and making it more certain to be observed in those localities. However, this also prevents realizing which harmonic of frequency most of the particles in that locality belong to [14]. As one to one correspondence exists between harmonics and momentums, the shorter the wavelength, the larger this change is in momentum. So, this superposition in a highly localized position makes it harder to realize which momentum is more significant in the region of interest (ROI). Because it gets harder and more uncertain to realize a very low wavelength component of a small region than of a wider one. Vice versa, as concerning ROI widens, one can have a more clear insight into constitutive harmonic components and less insight into their localizations [14].

## 2.2 Classical UP in Fourier analysis and spectral estimation

Any continuous function could be produced as an infinite summation of sine and cosine signals. The function that transforms a signal to the frequency domain is called the Fourier transform. It describes the signal intensity of each frequency component. The Fourier transform function is explained by the following formula:

$$Fourier\{g(t)\} = G(f) = \int_{-\infty}^{\infty} g(t) e^{-2\pi i f t} dt \qquad (2)$$

where *g(t)* is a continuous-time function, $\pi$ is the half of a circle perimeter with radius one and *G(f)* is extracted function in the frequency domain. While dealing with discrete data, plenty of issues take place. First of all, the sampling rate must be such that sufficient frequency information endures. Secondly, real data signals have limited durations which makes it necessary to use spectral estimation tools to estimate frequency characteristics. So, more localized information from the time domain makes it more uncertain to ensure about main harmonic components. Also, having certainty in a reduced region of frequencies makes it harder to judge the small region of signal behavior in the time domain. To prove this statement mathematically, one has to prove the following formula, stating that spectral and temporal information about a function cannot both sharply localized.

$$\forall g \in \varphi(\Box), \forall t_0 \in \Box, \forall f_0 \in \Box, \ \|G\| \times \|g\| = \|G\|_2^2 = \|g\|_2^2 \le 4\pi \|(t - t_0) g(t)\|_2 \|(f - f_0) \mathrm{G(f)}\|_2 \qquad (3)$$

Where *g(t)* is the corresponding function in time domain and *G(f)* is a function in frequency domain. ||g|| is the second norm of function *g(t)* (i.e. the integration of function square over time). Parseval theorem is symmetric w.r.t. time or frequency domain [16]. On the right side of the inequality, the first norm is a variance of information in the time domain and the second norm is the variance of spectral information in the Fourier domain. The higher the variance, the less centralized the corresponding domain. Proof of the stated inequality is through injecting Schwarz inequality and the Parseval theorem formula into the Fourier equation. For a mathematical proof of this statement, the reader can refer to [13] and [14]. Setting $t_0$ to $\|tg(t)\|_1$ and $f_0$ to $\|fg(f)\|_1$ turns rightmost norms to variance. Division of right side formula by most left side formula finally turns variances into standard deviation. It yields to classical form of the UP.

The subject of spectral estimation is clarified by listening to an instrument's sound. The higher duration the listened sound has, the more accurate the pitch will be estimated. The Brain cannot accurately recognize the tone frequency when the sound duration is low. For the example of listening to a piano, rapid pushing and leaving a piano button, makes it hard to discern which tone is played [14].

## 2.3 UP in the wavelet filter design

Spectral estimation states that, the more the data are localized in time by a window to increase time certainty, the less certain the truth of spectral information in the frequency domain gets. Wavelet is a set of spatiospectral filters that estimate spectral information dependent on temporal information in the least possible uncertain way. Precisely, the frequency resolution is multiple of temporal resolution per filter. The objective of this section is to only introduce concepts relevant to UP. For more information about wavelet design, the reader can refer to [15].

Not only is frequency analysis crucial for both small and wide temporal duration, but also this necessity can be generalized to the concepts of local and global optimizations. Frequency analysis in small time duration is insightful to know about temporary frequency distributions and harmonics. The certainty of harmonic estimations of low duration signals is reasonably low because of the UP [15]. As the time duration increases, the certainty of frequency estimation increases for a higher frequency resolution. In most engineering optimization problems, the global solution search can lead to a new convex environment and eventually new localities. Suitable global search results in a more reliable effective local searching process. A wider exploration space decreases fitness certainty of the sought solutions. On the other hand, when the searching process takes place in a more localized area, finding local optimums is more certain. As a result, both partly and holistic information are crucial to finding the optimum solution.

## 2.4 Extension of UP to a metaheuristic optimization

For implementing UP in metaheuristics, a set of roles should be extracted common to all applications. First, all mentioned applications deal with two variables that cannot be measured simultaneously with higher amounts of certainty. This characteristic is also evident in global searching versus local searching in an optimization problem. For the case of quantum mechanics, the wave function only describes the likelihood of lying in a specific position or momentum. Moreover, due to the unknown distribution of solutions, the probabilistic view can make much sense in an optimization problem. Another similarity is that each application has derived uncertainty inequality in a completely logical way. Similarly, in an optimization problem, the more localized the search takes place, the less globalized the search will be. Due to the high accuracy of local search and high consistency of global searching, both types of search can be useful. If an algorithm is capable of rightly deciding the extent of locality in iteration, a huge number of function evaluations can be saved.

In the next section, the main framework is proposed. This framework is based on the metrics capable of choosing regions in search space which provide new solutions in a more certain way according to currently available cost information. After the algorithm proposal, a set of useful metrics for attaining this objective is discussed.

# 3 UP based optimization (UPBO), the proposed approach

It is not necessary to know the concepts of the UP to implement or use the proposed optimization algorithm or improve it though it is inspired by the main concept of the Heizenburg UP. What is necessary to know, is the possibility of being inspired by newly added solutions, choose certain and suitable search space and guaranty about the betterment of solutions in one region over others. It is possible to adaptively tune the extent of space locality based on which solution update takes place. The concept of certainty in the optimization problem is defined by a metric being able to discern the probability that one region can lead to a better solution in itself. So, metrics are needed to be capable of comparing certainties.

The more local the search takes place, the more certain it is to find perfect solution for that locality; and simultaneously the less certain one can be that perfect global solution (for whole search space) is sought. Hence, an appropriate adjustment of the degree of locality for the selected search space is necessary. To be more simple, local and global search are both necessary and there should be a suitable metric for the choice of more certain solution spaces among various spaces with different volumes in which the solutions should update. So, the necessity of a metric to compare and decide between big search space and small search space has been felt here.

## 3.1 The general algorithm of the proposed framework

Supposing that measurement of a new solution in a local search region affects states of an imaginary quantity which is the best global optimization solution so far found. This effect has a certainty degree in the possibility of being the solution for a small region. By setting "the possibility of being the best solution for bigger regions" as another quantity, the measurement is less certain for this quantity respecting the first. Algorithm of the proposed framework is as follows:

**Input:**

Maximum number of iterations, preferred uncertainty metric, preferred convex hull type,
Maximum number of solution updates per iteration, Solution counts in popoulation

**Output:**

Least cost solution

**Process:**

Randomly generate n solutions in search space
While iteration< max_iter, do

- create n convex hulls with preferred kind ( convex hull methods proposed in table (1)) by randomly selecting hull centers from search space and using random radius.
- Compute preferred Uncertainty Metric for each convex hull using solutions existing in each cluster. (Uncertainty metric is shown in table (2))
- depending on the preferred metric, compute certainty of each cluster
- choose hulls with best certainties
- for each selected cluster
  - Generate $N_s*c_i/C_{all}$ solution using solution generating metric (table 3) in the selected cluster where $N_s$ is maximum number of solution updates per iteration, $C_{all}$ is sum of all certainties and $p(c_i)$ is certainty metric…
- Remove excessive worst costs and solutions to keep population up to parameter solutions_cnt

Figure 1 shows a schematic of the framework. Black dots indicate the solutions which their costs already exist in the memory. The bigger the dots get, the lower their costs get. Moreover, each hull's radius and center is selected randomly.

In this algorithm, fitnesses are attained by just a constant linear projection and translation of costs. Whenever costs are positive, they are subtracted from their maximum value. Otherwise, their minimum value are subtracted from each of them and results are saved as fitnesses.

Deciding when to explore or exploit is feasible through comparing the certainty of exploitation between big or small regions. For example, one case for certainty metric could be the density of average fitness. Exploration over exploitation is reasonably profitable when the smaller region fails in the density of average fitness over a larger region. The proposed certainty metrics are not necessarily the best and can be improved. However, after a series of experiments on the benchmark function used in the experimental phase, the preferred metric is chosen below. More comprehensive information about the algorithm is available in Figure 2 flowchart.

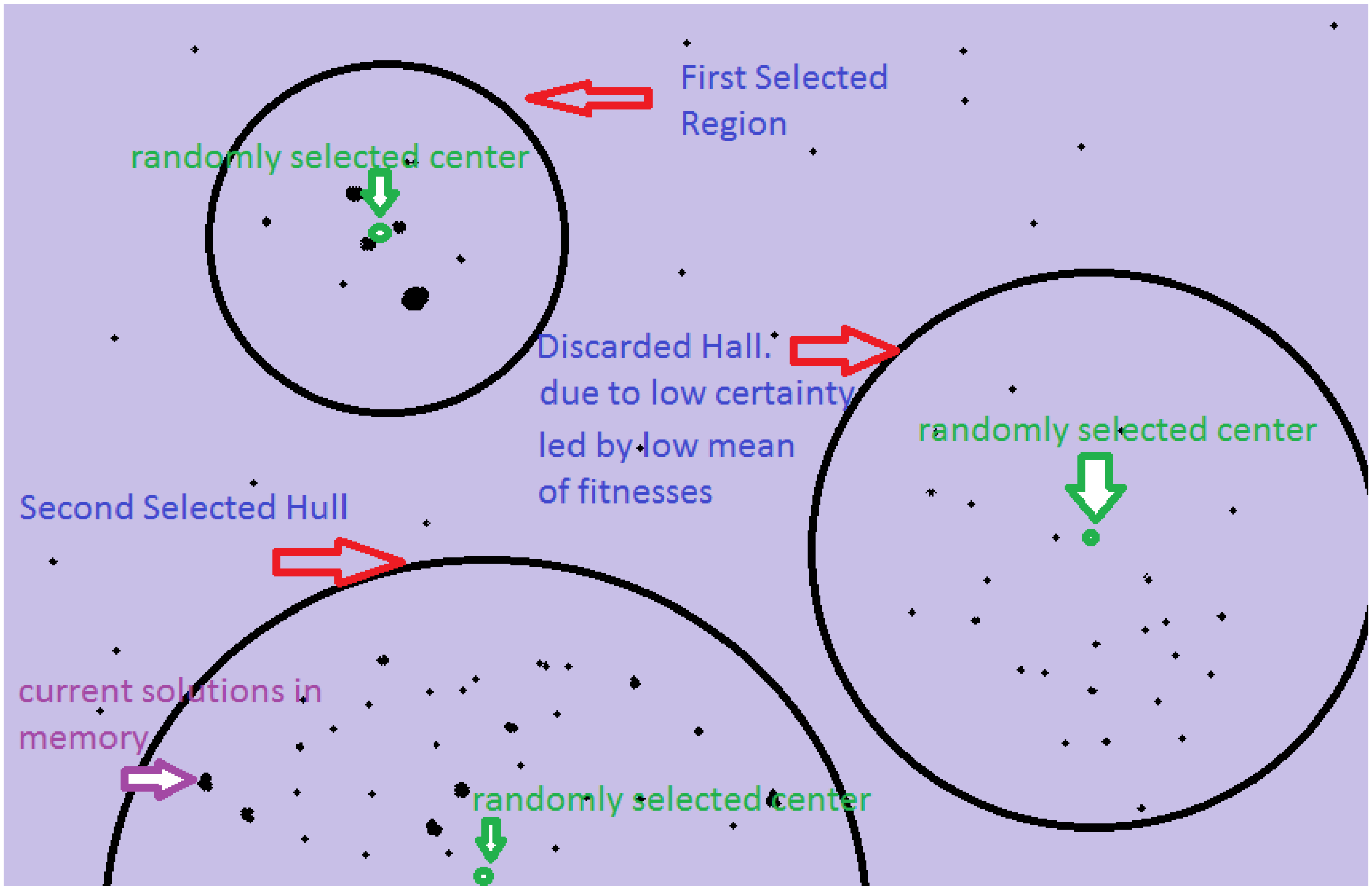

Figure 1; schematic of UPBO framework.

Any kind of metric capable of comparing certainty among different ROIs, more efficiently exploits currently available costs in search space. Usage of the term certainty is in the sense that sought solutions in a selected ROI give clue about better solutions in that region. Table 2 provides scale-comparable metrics capable of comparing ROIs in terms of different scales and locations. The proposed metrics are able to adaptively choose a large region over a small search region or vice versa. The choice of larger regions leads to global searching and more exploration while selecting a cluster with smaller space results in more exploitations.

## 3.2 Metric tables

For finding more smooth and convex localities in search space, guesstimating a suitable metric using the density of elites in local regions may be helpful. A metric with more certainty in local searching is more desired. Some metrics are needed to compare regions with different volumes and elites density to choose the best ROI for local searching. If the proposed metrics are appropriate, the finalized region will lead to a better solution. The objective is to propose metrics capable of comparing the certainty of selecting solutions from a variety of localities. Mean fitness per volume can be a suitable metric for comparing uncertainties. When fitness density is higher for the bigger area than of smaller one, solution selection in the larger area leads to a better result than the smaller one. This approach also removes the line between local and global search, proposing a more fuzzified approach for search space traversal.

Two tables for metrics are shown below. One is for convex hulls selecting localities, and the other is for different types of certainty metering which they are able to compare certainties for the existence of global solution among different regions and volumes. Sensitivity analysis and metrics comparison is conducted and the metric with best mean results for 50 independent tests have been set as default uncertainty metric. Results are shown in Table 3 and selected metrics are mentioned in the next section.

### 3.2.1 Metrics of convex hulls

Currently, only two types of the convex hull have been proposed which are detailed in Table 1. Results over 'Sphere' showed better performance in term of the sum of mean error benchmark functions than Sphere. During each iteration, solutions lying in multiple convex regions get selected using a preferred convex hull.

Table 1 - Appropriate Convex Hulls

| Convex Hull type | Suitability | Solution selection method |
|---|---|---|
| Sphere | Easy derivation of entailed solutions | By checking Euclidean distance to the center of the sphere |
| Ellipsoid | Better fitting to groups of solutions | Adding distances of a solution to both centers of ellipsoid should not exceed a threshold |

### 3.2.2 Certainty metrics

Suitable certainty metrics for various search ROIs are shown in table 2. Statistical analyses confirm the proposed metrics are capable of comparing hulls without variability to hull volumes and locations.

Table 2 - certainty metrics

| | Metric Tag in the algorithm | Description | Suitability |
|---|---|---|---|
| **The overall sum of all fitnesses in concerning cluster** | 'SumFitnessPerVolume' | Adds all fitness values in concerning hull. | The high amount of fitness values in a region of search space, can be a sign of the closeness of the region to a global solution. |
| **Sum of best solution fitness per volume** | 'SumEliteFitnessPerVol ume' | Finds n best solutions in volume in term of fitness and adds their values. | Instead of counting all fitnesses in computing uncertainty, only top solutions are selected. |
| **Average of Fitness values per each volume** | 'MeanFitnessPerVolume ' | Averages all fitness values in concerning hull. | To remove the negative effect of various volume size comparison, the average is used instead of sum for calculation of hull uncertainty. |
| **Best fitness value per volume** | 'BestFitnessPerVolume' | Find the best solution in hull in term of fitness and set fitness value a metric output. | Best solution fitness in each volume with any size can be a sign of certainty of locating a global solution in that volume. |
| **The variance of fitness values per volume** | 'VarFitnessPerVolume' | The variance of fitness values in concerning volume | In a hull where the spread of fitness values are high, there may be more likelihood of finding a better solution versus a hull with low fitness variance. |

The performance of each metric is evaluated in Table 3 which compares them with each other in terms of average function error (deviation of estimated global minimum from true minimum) over the corresponding group of benchmark functions. Functions described in Table (5) are divided into 3 subgroups of unimodal, multimodal and shifted/rotated functions.The first group checks out metric performance in local searching. On the other hands, second one checks out cooperation of explorations and exploitation. At last, the third one assays algorithm robustness led out of the metric. The results are already averaged out of 50 independent experiments of UPBO optimizer with sphere metric as hull and solution update method (Table (4)) with tag 'EitherRandomlyOrThroughBest'. Algorithm preferences have been cleared in Table (6).

Table 3- Comparison of each certainty metrics

| | **SumFitness Per Volume** | **SumEliteFitness Per Volume** | **MeanFitness Per Volume** | **BestFitness Per Volume** | **VarFitness Per Volume** |
|---|---|---|---|---|---|
| **unimodal** | 0.01667193 | 0.01640519 | 0.017513075 | 0.019160605 | 0.019259309 |
| **rank** | 2 | 1 | 3 | 4 | 5 |
| **multimodal** | 7.10177E-05 | 6.68016E-05 | 7.31606E-05 | 7.96947E-05 | 8.17945E-05 |
| **rank** | 3 | 2 | 1 | 4 | 5 |
| **shifted rotated functions** | 0.010920876 | 0.010359065 | 0.009618642 | 0.012451766 | 0.012055423 |
| **rank** | 3 | 2 | 1 | 5 | 4 |

Table 3 results suggest the metric of 'MeanFitnessPerVolume' for multimodal function optimizations. This rank went even higher than 'SumEliteFitnessPerVolume' metric and this is probably because some may not be a scale-invariant metric for comparing the certainty of two regions with different volumes. The role changes for unimodal functions and this is because of the concentrated distribution of high fitness solutions near a specific region. Overcoming 'SumEliteFitnessPerVolume' versus 'SumFitnessPerVolume' in both unimodal and multimodal functions shows the ineffectiveness of low fitness solutions in solution decisions. The fact that 'VarFitnessPerVolume' and 'BestFitnessPerVolume' got the lowest ranks over all groups, shows that neither spread of solutions nor merely best solution location can give sufficient information about clusters certainty in comparison view. Metric 'MeanFitnessPerVolume' and 'BestFitnessPerVolume' for shifted/rotated benchmark groups have ranked best and worst of all, respectively.

## 3.2.3 Solution Update methods

Table 4 describes solution update heuristics for finalized selected hulls during each iteration. For evaluating the performance of each approach, the mentioned benchmark functions in Table 5 are used and certainty metric with the tag of 'MeanFitnessPerVolume' is used. Each benchmark is tested out 50 times and specifications in *Table 7* is implemented. Relative rank of each method in terms of average function error (deviation of attained global minimum from true minimum) over the corresponding benchmark group (unimodal or multimodal) is evaluated.

Table 4- Solution Update methods

| Description | Method Tag | Description/ Pseudocode | Suitability | Relative Rank for unimodal functions | Relative rank for multimodal functions |
|---|---|---|---|---|---|
| The overall sum of all fitnesses in concerning cluster | 'EitherRandomlyOrThroughBest' | If a generated normalized uniform random number lower than predefined number 0<p<1, select solution randomly in the region using a uniform random vector generator within the intended hull. Otherwise, $X_{new}= X_{start} + u * (X_{best}- X_{start})$ where $X_{start}$ is an arbitrary select solution in concerning hull, and $X_{best}$ is the best solution in the intended hull. u is a uniform generated a random number between zero and one. | Random searching and structured searching gets combined. Tuning p during the run will give more flexibility to the algorithm. | 1 | 1 |
| Select new solution near in sequel of best solution of the hull and other selected one. | 'MoveThroughBest' | $X_{new}= X_{start} + u * (X_{best}- X_{start})$ where $X_{start}$ is an arbitrary select solution in concerning hull, and $X_{best}$ is the best solution in intended hull. u is a uniform generated a random number between zero and one. | Best currently sought solution may b closer to a best global solution. | 4 | 6 |
| Select two solutions and choose a new solution in the sequel. | 'Select2Sols&ChooseOneBetween' | Randomly choose two solutions $X_1$ and $X_2$ and set $X_{new}$= aX1+(1-a)X2=X2+(X1-X2)*a | Low fitness solutions get the filtered end of each iteration. By choosing a solution between two existing solutions, fitness improvement may be possible. | 7 | 7 |
| Mean of solution locations in cluster. | 'ClusterMean' | Insert new solution as mean of solutions contained in hull. | N/A | 6 | 5 |
| mean of N best solutions | 'MeanOfElites' | Set new solution as contained in hull. | Mean of elites locations may give a suitable estimate of the better unsought solution. | 2 | 2 |
| the weighted mean of solutions | 'GetWeightedMeanOfSols' | Insert new solution as the weighted mean of solutions contained in hull. Weights of each solution location, are their own witnesses. | The new solution will be selected near better solutions of the hull. It differs with MeanOfElites case in the fact that non-elite solutions will affect the selection of location. | 5 | 3 |
| N best solutions | 'GetWeightedMeanOfElites' | Insert new solution as a weighted mean of N best solutions contained in hull. Weights of selected solution location, are their own fitnesses. | In comparison with MeanOfElites case, The better the solution, the closer the new solution gets to. | 3 | 4 |

## 3.3 Proposed flowchart using preferred metrics

Due to ranks attained in the previous section, the finalized algorithm asserts that UPBO uses methods ‘Sphere’, ‘MeanFitnessPerVolume’ and 'EitherRandomlyOrThroughBest' respectively for hull type, certainty and solution update. The flowchart of the proposed algorithm is presented in figure 2.

# 4 Experimental Analysis

Benchmark function specifications are shown in Table 5 and Table *7*. Also, optimizer preferences are cleared up in Table 8.

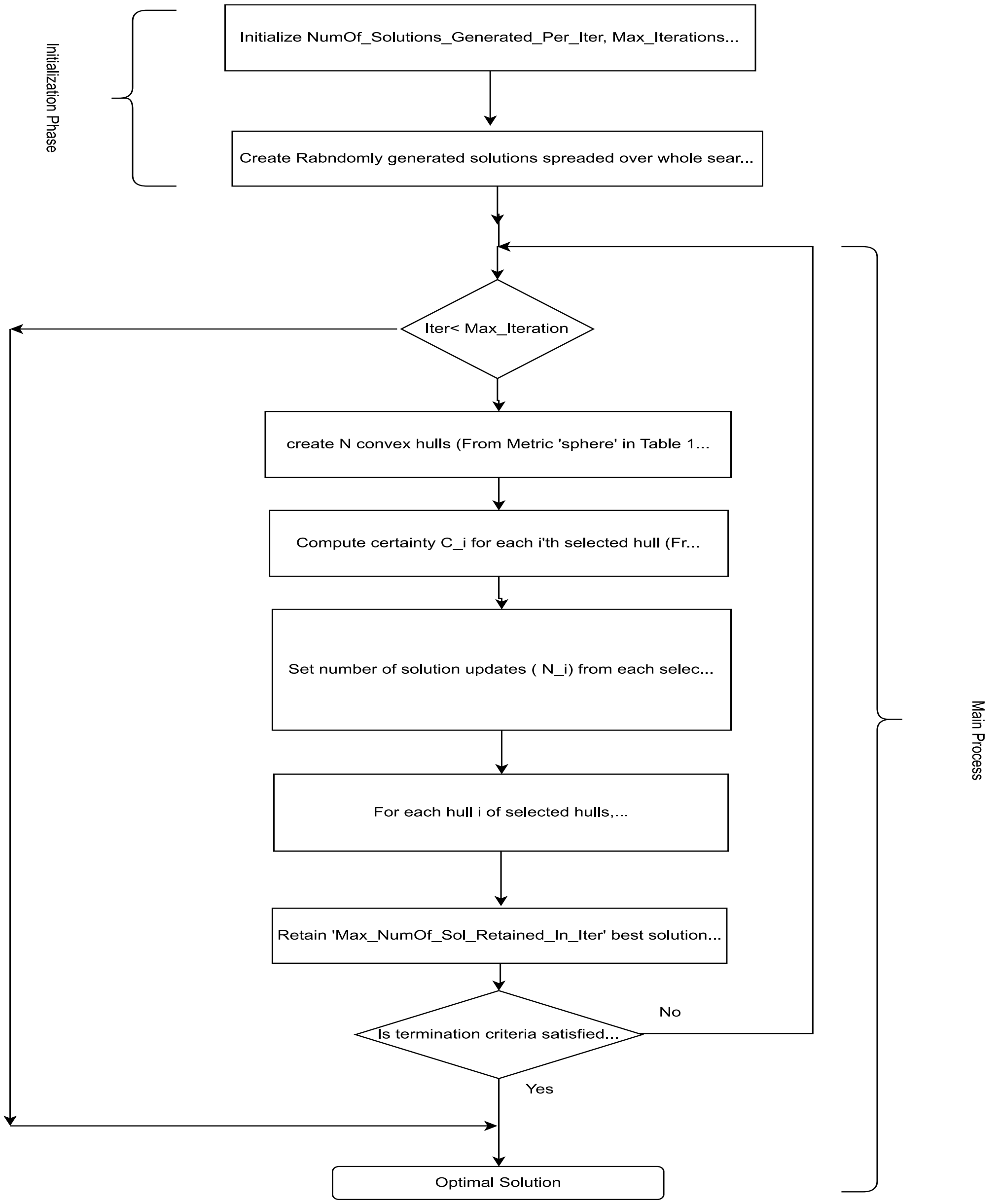

Figure 2- Flowchart of proposed UPBO

Table 5 is about benchmark functions used to compare optimizers. These functions have been selected from CEC2010 benchmarks in order of type. Four Types are selected in which, the first two are shown in this table and the other two types are demonstrated in Table 6. Types are respectively unimodal functions from F1 to F6 and multimodal ones ranged from F7 to F11. In Table 6, F12 to F16 are rotated functions and F17 to F20 are shifted rotated functions which are both for stability testing purposes. In upcoming results, types are shown in their own color to be discerned more easily.

Table 6- Benchmark functions used to compare optimizers

| Function | Range | Desired Optima | Formula | Surface Plot |
|---|---|---|---|---|
| **F1)Rosenbrock** | $x_i \in [-100,100], i = 1,\dots,d$ | Min = $\begin{cases} n=2, f(1,1)=0 \\ n=3, f(1,1,1)=0 \\ n>3, f(1_{1,1_2},\dots,1_n)=0 \end{cases}$ | $f_1(x)=\sum_{i=1}^{N-1} 100(x_{i+1}x_i^2)^2+(x_i)^2$ | 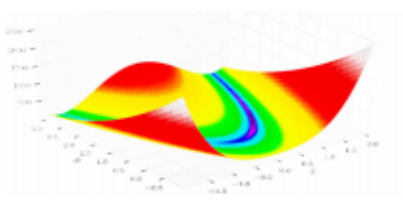 |
| **F2)Sphere** | $x_i \in [-5.12,5.12]$ | $f(x_1,\dots,x_n)=f(0,\dots,0)=0$ | $f_2(x)=\sum_{i=1}^{n} x_i^2$ | 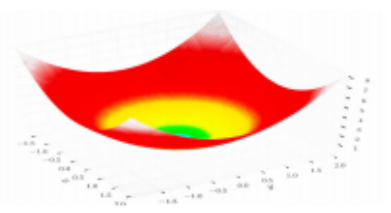 |
| **F3)Dixon Price** | $x_i \in [-10,10], i = 1,2$ | $f(x*)=0, \quad X_i = 2^{-\frac{2^i-2}{2^i}}, i=1,\dots,d$ | $f_3(x)=(x_1-1)^2+\sum_{i=2}^{d} i(2x_i^2 - x_{i-1})^2$ | 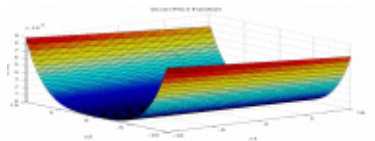 |
| **F4)Beale** | $x_i \in [-4.5,4.5], i = 1,2$ | $f(3,0.5)=0$ | $f_4(x,y)=(1.5-x+xy)^2+(2.25x+xy^2)^2$ | 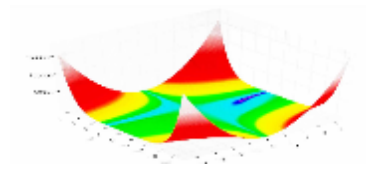 |
| **F5)Easom** | $x_i \in [-100,100], i = 1,\dots,d$ | $f(0,\dots,0)=0$ | $f_5(x) = -\cos(x_1)\cos(x_2)e^{-(x_1-\pi)^2-(x_2-\pi)^2}$ | 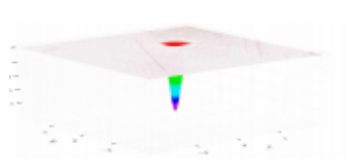 |
| **F6)Quartic** | $x_i \in [-1.28,1.28], i = 1,\dots,d$ | $f(0,\dots,0)=0$ | $f_{6(X)}=\sum_{i=1}^{n} ix_i^4$ | 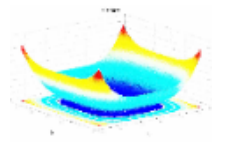 |
| **F7)Schwefel** | x_i $\in [-0.5,0.5], i = 1,\dots,d$ | $f(420.9687,\dots,420.9687)=0$ | $f_7(x) = 418.9829d-\sum_{i=1}^{d} x_i \sin\left(\sqrt{\lvert x_i \rvert}\right)$ | 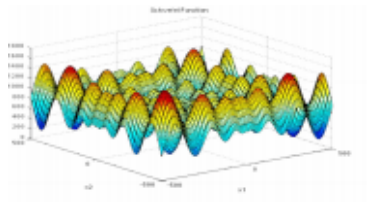 |
| **F8)Weierstrass** | $x_i \in [-0.5,0.5], i = 1,\dots,d$ | $f(x_{opt})=f(o)=x_{bias}=-0.5$ | $f_{8(x)}=\sum_{i=1}^{D}\left(\sum_{k=0}^{k_{max}} a^k\cos\left(2\pi b^{k(x_i+0.5)}\right)\right)-D.\sum_{k=0}^{k_{max}} a^k\cos(2\pi b^k.0.5)$ | 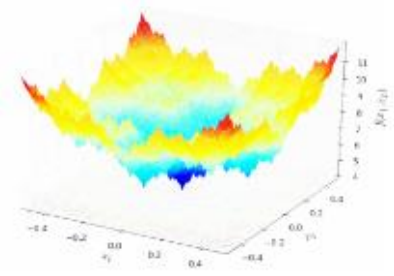 |
| **F9)Rastrigin** | $x_i \in [-5.12,5.12], i = 1,\dots,d.$ | $f(0,\dots,0)=0$<br>$a=0.5, b=3. K_{max}=20$ | $f_{9(x)}=\sum_{i=1}^{N}(x_i^2 10\cos(2\pi ix)+10)$ | 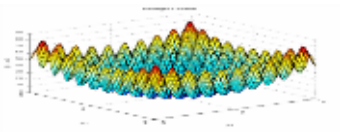 |
| **F10)Ackley** | $x_i \in [-32.768,32.768], i = 1,\dots,d$ | $f(0,\dots,0)=0$<br>$a=20, b=0.2$ and $c=2\pi$ | $f_{10(x)}=-ae^{-b\sqrt{\frac{1}{d}\times\sum_{i=1}^{d} x_i^2}} e^{\frac{1}{d}\times\sum_{i=1}^{d}\cos(cx_i)}+a+e$ | 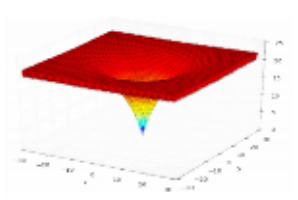 |
| **F11)Griewank** | $x_i \in [-600,600], i = 1,\dots,d.$ | $f(0,\dots,0)=0$ | $f_{11(x)}=\sum_{i=1}^{D}\frac{x_i^2}{4000}-\prod_{i=1}^{D}\cos\left(\frac{x_i}{\sqrt{i}}\right)$ | 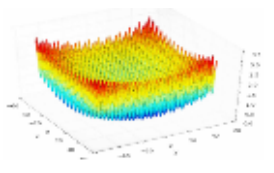 |

Table 7- Stability testing benchmark functions; Continuation of Table (5). F12 to F16 are rotated functions and F17 to F20 are shifted rotated functions which are both for stability testing purposes. In upcoming results, Types are shown in their own color to be discerned more easily.

| Function | Range | Desired Optima | Formula | Surface Plot |
|---|---|---|---|---|
| **F12)Rotated Ackley** | $xi \in [-32.768, 32.768], i = 1, \dots, d$ | $f(0,\dots,0) = 0$ | $f_{12(x)} = -20.e^{-0.2\sqrt{\frac{1}{D}\times\sum_{i=1}^{D} z_i^2}} e^{\frac{1}{D}\sum_{i=1}^{D}\cos(2\pi z_i)} + 20 + e + f_{opt}$ | 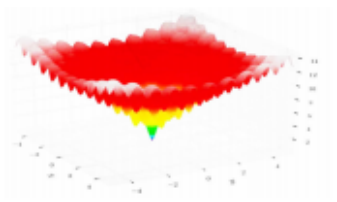 |
| **F13)Rotated Rastrigin** | $xi \in [-5.12, 5.12], i = 1, \dots, d.$ $z = R(0.0512.(x - x_(opt))$ | $f(0,\dots,0) = 0$ | $f_{13}(x) = \sum_{i=1}^{D} (z_i^2 10\cos(2\pi z_i) + 10) + f_{opt}$ | 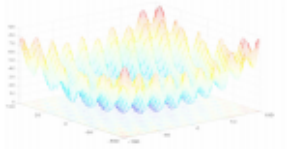 |
| **F14)Rotated Schwefel** | $xi \in [-500, 500], i = 1, \dots, d.$ | $f(x) = 0, x = \frac{1}{6} \times R^{-1} \times [420.9687, \dots, 420.9687]^T$ | $f_{14(x)} = 418.9829d - \sum_{i=1}^{d} z_i \sin\left(\sqrt{\lvert z_i \rvert}\right), z = R(6 \times x)$ | 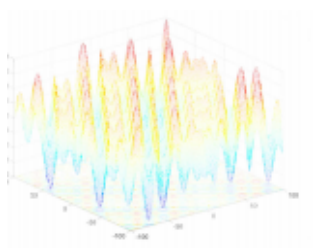 |
| **F15)Rotated Griewank** | $xi \in [-600, 600], i = 1, \dots, d.$ | $f(0,\dots,0) = 0$ | $f_{15(x)} = \sum_{i=1}^{D} \frac{z_i^2}{4000} - \prod_{i=1}^{D} \cos\left(\frac{z_i}{\sqrt{i}}\right) + 1 + f_{opt}$ , $z = R(6.x)$ | 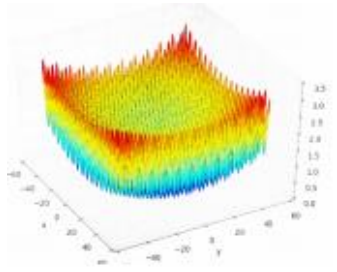 |
| **F16)Rotated Weierstrass** | $xi \in [-0.5, 0.5]$ $a = 0.5, b = 3. K_max = 20, z = R(0.005.x)$ | $f(0,\dots,0) = 0, sourcefunction$ | $\sum_{i=1}^{D}\left(\sum_{k=0}^{k_{max}} a^k \cos(2\pi b^{k(z_i+0.5)})\right) - D.\sum_{k=0}^{k_{max}} a^k \cos(2\pi b^k.0.5) + f_{opt}$ | 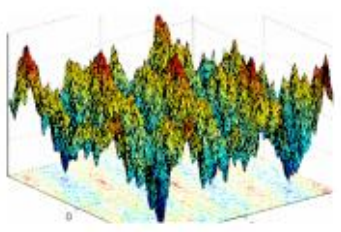 |
| **F17)Rotate Shift Expand Scaffer** | $xi \in [-100, 100]$ | $f(0,\dots,0) = 0, sourcefution$ | $\sum_{i=1}^{D-1} p(z_i, z_{i+1}) + p(z_D, z_1)$ $p(u,y) = \frac{\sin\left(\sqrt{u^2+y^2}\right) - 0.5}{(1 + 0.001.(u^2+y^2))^2} + 0.5$ | 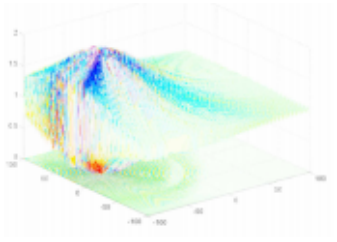 |
| **F18)Rotate Shift Griewank** | $xi \in [-600, 600], i = 1, \dots, d.$ | $f(0,\dots,0) = 0, sourcefunction$ | $\sum_{i=1}^{D} \frac{z_i^2}{4000} - \prod_{i=1}^{D} \cos\left(\frac{z_i}{\sqrt{i}}\right)$ z=R(6.(x-x_opt)) | 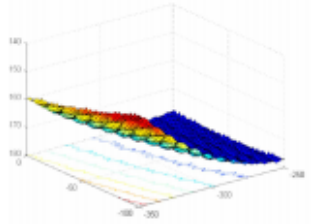 |
| **F19)Rotate Shift Rastrigrin** | $xi \in [-5.12, 5.12], i = 1, \dots, d.$ | $f(0,\dots,0) = 0, sourcefunction$ | $f_{19(x)} = \sum_{i=1}^{D} (z_i^2 10\cos(2\pi z_i) + 10) + f_{(bias)_{10}}$ | 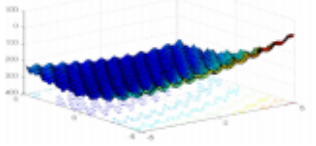 |
| **F20)Rotate Shift Ackly** | $x_i \in [-32.768, 32.768], i = 1, \dots, d$ | $f(0,\dots,0) = 0, sourcefunction$ | $f_{20(x)} = -20e^{-0.2\sqrt{\frac{1}{D}\times\sum_{i=1}^{D} z_i^2}} e^{\frac{1}{d}\times\sum_{i=1}^{d}\cos(2\pi z_i)} + 20 + e + f_(bias)$ | 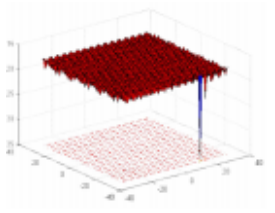 |

Table 8- Optimizers' preferences.

| **CLPSO [24]** | **ICA [25]** | **CICA [26]** | **GA** | **SA** | **BA [27]** | **PSO [28]** | **PSO-W** | **PSO-w-local** | **UPBO*** |
|---|---|---|---|---|---|---|---|---|---|
| Inertia= [0.7,0.9] | Number of Contries: 60 | Number of Contries: 60 | Mutation: 1% | $T_{min}$=1 | $V_i$: randomly uniform | Inertia= [0.7,0.9] | Inertia= [0.7,0.9] | Inertia= [0.7,0.9] | NumOfHulls= 7 |
| Speed limit: 0.1/8*[$X_{min}$ , $X_{max}$] | Imperialists=10 | Imperialists=10 | Crossover: 0.8 | $T_{max}$=$10^{-4}$ | $F_i$=0.5 | Speed limit: 0.1/8[$X_{min}$ , $X_{max}$] | Speed limit: 0.1/8[$X_{min}$ , $X_{max}$] | Speed limit: 0.1/8[$X_{min}$ , $X_{max}$] | HullType='sphere' CertaintyMetricType='MeanFitnessPerVolume' UpdateClusterSolutionMethod='EitherRandomlyOrThroughBest' |
| | Possessed colonies = 120 | Possessed colonies = 120 | Replacement: 50% | Repeats per state: 1,2,4,… .4096 | | | | | SpheresRadiusMin=0.2 SpheresRadiusMax=0.7 |
| | | | | | | | | | EliteCostsThresh=0.3 |

Parameter 'EliteCostsThresh'=0.3 is used in metrics containing the word 'Elites' in their tags. For the selection of elites, only those solutions are selected whose costs are less than 0.3 of all mean costs in the respective hull. In order to discover the performance of the algorithm, it is compared with other well-known metaheuristics in terms of average error of global optima location from its true position. Average has been taken place among 50 trials. A paired T-test has been used to assess whether or not the mean results of an optimum error are reliable. The null hypothesis is the indifference of proposed UPBO concerning evaluating optimizer per benchmark function. P-value results are brought in Table 10. Results show the significance of experiments in rejecting the null hypothesis due to the fact that p-values are less than 0.05.

Analysis of results is performed using three kinds of benchmark functions with well-known metaheuristic algorithms. Benchmark function consists of unimodal, multimodal and also shifted or rotated functions. The performance of this framework is assessed against the roots of well-known metaheuristics. Although further comparisons between its performance and recent methods have been done. But these recent algorithms are not the most powerful ones. Because still, this algorithm has weakly tuned parameters with weak uncertainty metrics and comparing it to the strongest methods can distort viewpoints to the proposed idea. Although comparisons over other optimizers have not been made, optimizers like ICA, CICA, and BA (bat algorithm) are also new.

Evaluations and analyses have shown the performance of this algorithm over most of the functions of benchmarks dictionary over mentioned evaluating optimizers. The outcome shows the power of UPBO not only in multimodal functions but also in rotated and shifted-rotated functions as a sign of robustness. As an algorithm handles one group of functions, it will manifest weakness in the other. By preventing the algorithm from falling into local optima, the degree of exploration has to get increased, making algorithms tackle multimodal functions more easily but show lags in unimodal function optimizations.

The experimental analysis demonstrates the performance of the UPBO on multimodal functions Beale, Ackley, Grievank and Rastrigin, and good performance on Easom and Rosenbrock. Other multimodal functions' performances are also comparable between UPBO and well-known algorithms. Also, in the fields of unimodal functions like sphere, booth, and Bohachevsky the proposed algorithm is outperforming baselines. For analysis of robustness, this algorithm manifests fairly good performance over shifted rotated multimodal functions like Ackley, Rastrigin and Grievank and Scaffer functions which have fast changes in their localities.

In Table 10, outperformance per function is shown for each algorithm over other algorithms, having overall number of error outperformance summed up. As that results are mean values of 50 independent experiments, they take decimal values. According to Table 10, UPBO outperforms all mentioned algorithms and moreover, it is comparable to powerful recent methods.

UPBO has shown weakness in performances of functions like Schwefel and Weierstrass over powerful algorithms. This shows UPBO is yet unripe for functions with fast changes. It is guesstimated that, by appropriate choice of cluster volume range, the problem can also be addressed to prevent algorithm from selecting very large hull areas. In upcoming supplementary works, estimation of the range of hulls volumes with help of function rate of change, will be put into the plan.

# 5 Conclusion

There have been a large number of researches proving that physics inspired metaheuristics have proved helpful in the brand-new engineering concepts like Context Aware Information Engineering, Edge Computing and communication networks, where a lot of physical laws interact all in a unified network. A new metaheuristic framework is proposed and its suitable metrics have been derived. The best-suited metric during each phase has been selected and the finalized algorithm out of framework is shown in the flowchart. The performance demonstrated that UPBO can be a good approach to deal with multimodal functions. It is hoped that deficiencies of the algorithm in unimodal function optimization get compensated. To attain that, we will propose more powerful methods and metrics for framework first in solution update and second in certainty metrics. Performance analysis has shown that the algorithm is able to compete with some of the powerful and recently proposed methods in metaheuristic optimization.

Table 9- All functions with NFE=30000 had 3 dimensional input vectors. Dimensions for 180000 and 500000 cases were 10 and 30 respectively.

| | | Type | NFE | CLPSO | ICA | CICA | GA | SA | BA | PSO | PSO-W | PSO-w-local | UPBO* |
|---|---|---|---|---|---|---|---|---|---|---|---|---|---|
| **F1** | **Rosenbrock** | **Unimodal** | 180000 | 2.46E+00 ± 1.70E+00 | 2.00E-01 ± 3.60E-01 | 2.00E-02 ± 2.00E-02 | 1.63E+01 ± 7.60E+00 | 2.09E-04 ± 2.67E-04 | 1.97E+01 ± 1.42E+01 | 2.14E+00 ± 1.81E+00 | 3.08E+00 ± 7.69E-01 | 3.92E+00 ± 1.19E+00 | **7.62E-16 ± 6.91E-17** |
| **F2** | **Sphere** | **Unimodal** | 180000 | 5.15E-29 ± 2.16E-28 | 2.87E+01 ± 2.07E+00 | 2.50E-07 ± 1.10E-01 | 3.92E-05 ± 4.86E-04 | 4.14E-03 ± 4.87E-03 | 7.85E-21 ± 3.74E-23 | 1.23E-30 ± 3.41E-29 | 9.78E-30 ± 2.50E-29 | **1.23E-30 ± 6.17E-30** | 1.49E-04 ± 1.24E-06 |
| **F3** | **Dixon Price** | **Unimodal** | 500000 | 1.32E-05 ± 6.95E-07 | 9.11E+00 ± 3.73E-01 | 6.28E-02 ± 5.07E-03 | 6.67E-01 ± 5.32E-02 | 1.71E+03 ± 6.92E+01 | 6.67E-01 ± 5.07E-02 | 1.00E+00 ± 5.05E-02 | 1.90E-02 ± 1.17E-03 | 3.40E-06 ± 2.30E-06 | 1.00E-01 ± 9.33E-04 |
| **F4** | **Beale** | **Unimodal** | 500000 | 9.04E-02 ± 9.79E-03 | 5.05E+02 ± 2.53E+00 | 9.40E-01 ± 6.32E-02 | 2.68E-11 ± 2.83E-12 | 4.60E-06 ± 4.27E-07 | 7.62E-01 ± 4.91E-02 | 1.42E+01 ± 1.28E+00 | 1.01E-01 ± 6.72E-03 | 9.93E-02 ± 1.59E-01 | **3.35E-15 ± 4.27E-14** |
| **F5** | **Easom** | **Unimodal** | 500000 | 6.52E-08 ± 9.39E-09 | 6.26E-11 ± 7.71E-12 | 3.27E-06 ± 4.29E-07 | -8.11E-05 ± -7.41E-06 | 1.23E-30 ± 9.66E-29 | **1.23E-30 ± 3.65E-29** | 2.68E-09 ± 3.34E-10 | 7.27E-07 ± 7.92E-08 | 7.05E-08 ± 3.40E-08 | 2.52E-13 ± 3.11E-15 |
| **F6** | **Quartic** | **Unimodal** | 500000 | **1.23E-30 ± 7.66E-24** | 1.23E-30 ± 1.25E-19 | 1.23E-30 ± 3.82E-29 | 1.23E-30 ± 1.70E-30 | 7.81E-12 ± 8.07E-13 | **1.23E-30 ± 4.28E-29** | 1.23E-30 ± 2.41E-29 | 1.23E-30 ± 8.61E-29 | 1.23E-30 ± 9.43E-30 | 6.01E-05 ± 4.03E-07 |
| **F7** | **Schwefel** | **Multimodal** | 180000 | **1.23E-30 ± 5.54E-29** | 5.99E-01 ± 6.66E-02 | 1.98E+04 ± 7.94E+01 | 1.08E+01 ± 2.08E-01 | 5.04E-04 ± 7.02E-05 | 2.56E+01 ± 2.36E+00 | 9.82E-02 ± 7.21E-03 | 3.20E+02 ± 1.63E+00 | 3.26E+02 ± 1.32E+02 | 3.61E-04 ± 5.07E-06 |
| **F8** | **Weierstrass** | **Multimodal** | 180000 | **1.23E-30 ± 6.27E-29** | 5.51E-02 ± 2.09E-01 | 1.29E-04 ± 6.60E-04 | 4.35E-05 ± 2.77E-05 | 3.31E-03 ± 2.74E-03 | 5.33E-12 ± 5.99E-08 | 1.23E-30 ± 9.06E-29 | 1.30E-04 ± 3.30E-04 | 1.41E-06 ± 6.31E-06 | 1.50E-15 ± 1.62E-17 |
| **F9** | **Rastrigin** | **Multimodal** | 500000 | **1.23E-30 ± 4.86E-29** | 1.66E-06 ± 9.12E-06 | 9.34E-09 ± 3.42E-08 | 4.75E-10 ± 2.14E-08 | 9.79E-05 ± 1.89E-03 | 7.96E+00 ± 8.61E+00 | 9.95E-01 ± 6.09E-01 | 6.76E+02 ± 5.43E+01 | 3.88E+00 ± 2.30E+00 | 4.81E-10 ± 2.78E-9 |
| **F10** | **Ackley** | **Multimodal** | 500000 | 4.32E-14 ± 2.55E-14 | 7.11E-05 ± 8.20E-06 | 1.02E-07 ± 1.23E-07 | 1.47E-05 ± 8.97E-08 | 5.04E-04 ± 4.23E-04 | 2.63E-12 ± 2.49E-12 | **1.23E-30 ± 7.33E-29** | 6.32E-11 ± 1.73E-15 | 6.04E-15 ± 1.67E-15 | **1.02E-30 ± 1.44E-30** |
| **F11** | **Griewank** | **Multimodal** | 500000 | 4.56E-03 ± 4.81E-03 | 1.03E-10 ± 8.14E-10 | 3.47E-14 ± 5.07E-15 | 1.56E+01 ± 2.08E+01 | 1.34E-03 ± 1.95E-04 | 1.36E-09 ± 2.71E-06 | 1.14E-01 ± 4.96E-02 | 2.43E-01 ± 3.07E-02 | 7.80E-02 ± 3.79E-02 | **8.08E-18 ± 2.01E-19** |
| **F12** | **Rotated Ackley** | **Robustness Evaluator** | 30000 | 3.56E-05 ± 4.35E-06 | 3.39E+02 ± 1.96E+00 | 1.92E+00 ± 2.41E-01 | 1.55E+01 ± 1.54E+00 | 2.08E-04 ± 1.62E-05 | 1.99E+01 ± 4.78E-01 | 1.73E+01 ± 3.43E-01 | 2.80E-01 ± 2.11E-02 | **6.39E-15 ± 3.18E-15** | 1.55E-04 ± 1.52E-06 |
| **F13** | **Rotated Rastrigin** | **Robustness Evaluator** | 30000 | 5.97E+00 ± 2.93E-01 | 2.48E+02 ± 4.94E+00 | 7.51E+01 ± 6.17E+00 | 1.54E+05 ± 4.48E+02 | 4.14E+00 ± 3.42E-01 | 4.48E+00 ± 1.89E-01 | 1.62E+01 ± 5.45E-01 | 9.90E+00 ± 4.87E-01 | 9.25E+00 ± 2.74E+00 | **3.76E-02 ± 3.69E-04** |
| **F14** | **Rotated Schwefel** | **Robustness Evaluator** | 30000 | 1.14E+02 ± 1.30E+00 | 2.32E+06 ± 8.09E+02 | 1.07E+04 ± 1.46E+01 | 3.62E+03 ± 1.33E+02 | 5.36E-02 ± 6.76E-03 | 3.84E+01 ± 2.20E+00 | 5.35E+03 ± 8.45E+01 | 5.69E+02 ± 3.66E+00 | 4.72E+02 ± 3.07E+02 | **5.35E-02 ± 4.71E-04** |
| **F15** | **Rotated Griewank** | **Robustness Evaluator** | 180000 | 4.50E-02 ± 4.50E-03 | 2.56E-02 ± 1.79E-03 | 2.68E+01 ± 6.97E-01 | 5.61E-06 ± 6.03E-07 | **3.36E-06 ± 3.54E-07** | 1.11E-15 ± 1.61E-16 | 1.44E-02 ± 8.86E-04 | 2.54E+00 ± 2.13E-01 | 8.04E-02 ± 4.46E-02 | 2.29E-04 ± 1.77E-06 |
| **F16** | **Rotated Weierstrass** | **Robustness Evaluator** | 180000 | 3.72E-10 ± 5.23E-11 | 6.83E-07 ± 3.55E-08 | 6.04E+01 ± 4.14E+00 | 1.22E-04 ± 1.56E-05 | 3.65E-07 ± 2.99E-08 | 9.54E-07 ± 7.08E-08 | 9.32E-06 ± 7.58E-07 | 3.74E+00 ± 2.13E-01 | 2.14E-01 ± 3.65E-01 | **9.54E-12 ± 5.92E-14** |
| **F17** | **Rotate Shift Expand Scaffer** | **Robustness Evaluator** | 30000 | 1.76E+00 ± 1.65E-01 | 4.82E+01 ± 2.18E+00 | 4.12E+01 ± 6.27E-01 | 4.88E+00 ± 1.80E-01 | 4.91E-05 ± 6.41E-06 | 4.51E+00 ± 3.95E-01 | 4.90E+00 ± 1.72E-01 | 2.42E+00 ± 1.92E-01 | 1.92E+00 ± 7.77E-01 | **4.89E-05 ± 3.05E-07** |
| **F18** | **Rotate Shift Griewank** | **Robustness Evaluator** | 30000 | 3.55E+01 ± 7.02E-01 | 2.74E+03 ± 3.80E+01 | 6.52E+02 ± 3.74E+00 | 4.34E+02 ± 5.00E+01 | 5.51E-03 ± 4.58E-04 | 2.19E-01 ± 1.18E-02 | 5.38E+01 ± 6.59E+00 | 4.26E+01 ± 8.80E-01 | 3.92E+01 ± 7.64E+00 | **1.63E-03 ± 1.02E-05** |
| **F19** | **Rotate Shift Rastrigrin** | **Robustness Evaluator** | 30000 | 6.01E+01 ± 1.78E+00 | 2.66E+04 ± 2.37E+01 | 2.41E+04 ± 1.81E+02 | 6.58E+02 ± 6.30E+01 | 9.32E-03 ± 1.08E-03 | 9.48E+02 ± 7.76E+00 | 6.92E+02 ± 2.57E+01 | 3.11E+02 ± 1.64E+01 | 1.68E+02 ± 2.63E+02 | **3.49E-04 ± 3.63E-06** |
| **F20** | **Rotate Shift Ackly** | **Robustness Evaluator** | 180000 | 7.20E+00 ± 4.15E-01 | 2.81E+02 ± 2.92E+00 | 1.24E+02 ± 2.11E+00 | 2.02E+01 ± 1.19E+00 | 2.15E-04 ± 2.46E-05 | 2.00E+01 ± 2.34E+00 | 2.07E+01 ± 2.76E+00 | 9.01E+00 ± 4.08E-01 | 7.23E+00 ± 4.99E+00 | **2.11E-04 ± 2.11E-06** |

Table 10- Number of algorithms outperformance from each other

| | **NFE** | **CLPSO** | **ICA** | **CICA** | **GA** | **SA** | **BA** | **PSO** | **PSO-W** | **PSO-W-LOCAL** | **UPBO** |
|---|---|---|---|---|---|---|---|---|---|---|---|
| **Sphere** | 180000 | 6 | 0 | 4 | 3 | 1 | 5 | **9** | 7 | **8** | 2 |
| **Rosenbrock** | 100000 | 4 | 6 | 7 | 1 | **8** | 0 | 5 | 3 | 2 | **9** |
| **Ackley** | 500000 | 6 | 1 | 3 | 2 | 0 | 5 | **8** | 4 | 7 | **9** |
| **Griewank** | 500000 | 4 | 7 | **8** | 0 | 5 | 6 | 2 | 1 | 3 | **9** |
| **Weierstrass** | 180000 | **8** | 0 | 3 | 4 | 1 | 6 | **9** | 2 | 5 | 7 |
| **Rastrigin** | 500000 | **8** | 5 | 6 | 7 | 4 | 1 | 3 | 0 | 2 | **9** |
| **Schwefel Rotated** | 100000 | **9** | 5 | 0 | 4 | 7 | 3 | 6 | 2 | 1 | **8** |
| **Ackley Rotated** | 30000 | **8** | 0 | 4 | 3 | 6 | 1 | 2 | 5 | 9 | 7 |
| **Griewank Rotated** | 180000 | 3 | 4 | 0 | 7 | **8** | 9 | 5 | 1 | 2 | 6 |
| **Weierstrass Rotated** | 180000 | **8** | 6 | 0 | 3 | 7 | 5 | 4 | 1 | 2 | **9** |
| **Rastrigin Rotated** | 30000 | 6 | 1 | 2 | 0 | **8** | 7 | 3 | 4 | 5 | **9** |
| **Schwefel** | 30000 | 6 | 0 | 1 | 3 | **8** | 7 | 2 | 4 | 5 | **9** |
| **Beale** | 500000 | 6 | 0 | 2 | **8** | 7 | 3 | 1 | 4 | 5 | **9** |
| **Easom** | 500000 | 3 | 5 | 0 | **9** | 7 | **8** | 4 | 1 | 2 | 6 |
| **Quartic Rotate Shift** | 500000 | 4 | 5 | 6 | 2 | 1 | 3 | 7 | **8** | **9** | 0 |
| **Ackly Rotate Shift Expand** | 180000 | 7 | 0 | 1 | 3 | **8** | 4 | 2 | 5 | 6 | **9** |
| **Scaffer Rotate Shift** | 30000 | 7 | 0 | 1 | 3 | **8** | 4 | 2 | 5 | 6 | **9** |
| **Griewank Rotate Shift** | 30000 | 6 | 0 | 1 | 2 | **8** | 7 | 3 | 4 | 5 | **9** |
| **Rastrigrin** | 30000 | 7 | 0 | 1 | 4 | **8** | 2 | 3 | 5 | 6 | **9** |
| **Dixon Price** | 500000 | **8** | 1 | 6 | 3 | 0 | 4 | 2 | 7 | **9** | 5 |
| **Total Score** | | 246 | 99 | 117 | 139 | 219 | 181 | 169 | 160 | 167 | **303** |

Table 11- Statistical Analysis results. Paired T-test statistical analysis of proposed algorithm versus each evaluator mentioned per column.

| *UPBO* | **CLPSO** | **ICA** | **CICA** | **GA** | **SA** | **BA** | **PSO** | **PSO-W** | **PSO-W-Local** |
|---|---|---|---|---|---|---|---|---|---|
| ***F1*** | 0.0000 | 0.0000 | 0.0000 | 0.0000 | 0.0000 | 0.0000 | 0.0000 | 0.0000 | 0.0000 |
| ***F2*** | 0.0000 | 0.0000 | 0.0000 | 0.0000 | 0.0000 | 0.0000 | 0.0000 | 0.0000 | 0.0000 |
| ***F3*** | 0.0000 | 0.0000 | 0.0000 | 0.0000 | 0.0000 | 0.0000 | **NA** | 0.0000 | 0.0000 |
| ***F4*** | 0.0000 | 0.0000 | 0.0000 | 0.0000 | 0.0000 | 0.0000 | 0.0000 | 0.0000 | 0.0000 |
| ***F5*** | 0.0000 | 0.0000 | 0.0000 | 0.0000 | 0.0000 | 0.0000 | 0.0000 | 0.0000 | 0.0000 |
| ***F6*** | **NA** | 0.0000 | 0.0000 | 0.0000 | 0.0000 | 0.0000 | 0.0000 | 0.0000 | 0.0000 |
| ***F7*** | 0.0000 | 0.0000 | 0.0000 | 0.0000 | 0.0000 | 0.0000 | 0.0000 | 0.0000 | 0.0000 |
| ***F8*** | 0.0000 | 0.0000 | 0.0000 | 0.0000 | 0.0000 | 0.0000 | 0.0000 | 0.0000 | 0.0000 |
| ***F9*** | 0.0000 | 0.0000 | 0.0000 | 0.0000 | 0.0000 | 0.0000 | 0.0000 | 0.0000 | 0.0000 |
| ***F10*** | 0.0000 | 0.0000 | 0.0000 | 0.0000 | 0.0000 | 0.0000 | 0.0000 | 0.0000 | 0.0000 |
| ***F11*** | 0.0000 | 0.0000 | 0.0000 | 0.0000 | 0.0000 | 0.0000 | 0.0000 | 0.0000 | 0.0000 |
| ***F12*** | 0.0000 | 0.0000 | 0.0000 | 0.0000 | **0.5265** | 0.0000 | 0.0000 | 0.0000 | 0.0000 |
| ***F13*** | 0.0000 | 0.0000 | 0.0000 | 0.0000 | 0.0000 | 0.0000 | 0.0000 | 0.0000 | 0.0000 |
| ***F14*** | 0.0000 | 0.0000 | 0.0000 | 0.0000 | 0.0000 | 0.0000 | 0.0000 | 0.0000 | 0.0000 |
| ***F15*** | 0.0000 | 0.0000 | 0.0000 | 0.0000 | 0.0000 | 0.0000 | 0.0000 | 0.0000 | 0.0000 |
| ***F16*** | 0.0000 | 0.0000 | 0.0000 | 0.0000 | 0.0000 | 0.0000 | 0.0000 | 0.0000 | 0.0000 |
| ***F17*** | 0.0000 | 0.0000 | 0.0000 | 0.0000 | 0.0001 | 0.0000 | 0.0000 | 0.0000 | 0.0000 |
| ***F18*** | 0.0000 | 0.0000 | 0.0000 | 0.0000 | 0.0000 | 0.0000 | 0.0000 | 0.0000 | 0.0000 |
| ***F19*** | 0.0000 | 0.0000 | 0.0000 | 0.0000 | 0.0000 | 0.0000 | 0.0000 | 0.0000 | 0.0000 |
| ***F20*** | 0.0000 | 0.0000 | 0.0000 | 0.0000 | 0.0000 | 0.0000 | 0.0000 | 0.0000 | 0.0000 |